%% file: main.tex
\documentclass[10pt,twocolumn,letterpaper]{article}

\usepackage{iccv}
\usepackage{times}
\usepackage{epsfig}
\usepackage{graphicx}
\usepackage{amsmath}
\usepackage{amssymb}
\usepackage{bbm}
\usepackage[usenames,dvipsnames]{xcolor}
\usepackage{pifont}
\usepackage{multirow}
\usepackage{enumitem}
 \usepackage{url}

\usepackage{xcolor}

\usepackage{nicefrac}
\definecolor{KleinBlue}{rgb}{0.0, 0.129, 0.7}
\definecolor{C1}{HTML}{c9734e}
\definecolor{C2}{HTML}{578eb4}
\definecolor{C3}{HTML}{7a71af}
\definecolor{C4}{HTML}{64a94f}
\definecolor{C5}{HTML}{345f92}

\newcommand{\greencheck}{\textcolor{OliveGreen}{\ding{51}}}
\newcommand{\redx}{\textcolor{Red}{\ding{55}}}

\usepackage[pagebackref=true,breaklinks=true,letterpaper=true,colorlinks,bookmarks=false]{hyperref}
\usepackage{wrapfig}
\usepackage{textcomp}
\hypersetup{
    colorlinks=true,
    linkcolor=blue,
    urlcolor=blue,citecolor=KleinBlue
}
\iccvfinalcopy %

\def\iccvPaperID{3507} %

\ificcvfinal\fi

\begin{document}

\title{ Rapid Adaptation in Online Continual Learning: Are We Evaluating It Right?}
\author{
\text{Hasan Abed Al Kader Hammoud}\textsuperscript{1*} \quad \text{Ameya Prabhu}\textsuperscript{2}\thanks{ authors contributed equally; order decided by a coin flip.} \quad \text{Ser-Nam Lim}\textsuperscript{3}\quad \text{Philip H.S. Torr}\textsuperscript{2} \\ \quad \text{Adel Bibi}\textsuperscript{2} \quad \text{Bernard Ghanem}\textsuperscript{1}\\
KAUST\textsuperscript{1} \quad University of Oxford\textsuperscript{2} \quad  \quad Meta AI\textsuperscript{3}}
\maketitle

\newcommand{\bibi}[1]{\textcolor{red}{bibi: #1}}
\newcommand{\adel}[1]{\todo[inline]{{\textbf{Bibi:} \emph{#1}}}}
\newcommand{\ameya}[1]{\todo[inline]{{\textbf{Ameya:} \emph{#1}}}}
\newcommand{\hasan}[1]{\todo[inline]{{\textbf{Hasan:} \emph{#1}}}}

\def\eg{\emph{e.g}\onedot} \def\ie{\emph{i.e}\onedot} 
\newcommand{\bibir}[1]{\textcolor{red}{{[\textbf{Bibi:} \emph{#1}]}}}
\ificcvfinal\thispagestyle{empty}\fi

\vspace*{-0.8cm}
\input{sections/abstract}
\input{sections/introduction}
\input{sections/relatedwork.tex}
\input{sections/our_methodology}

\input{sections/experiments}

\input{sections/conclusion}

\clearpage
{\small
\bibliographystyle{ieee_fullname}
\bibliography{main}
}

\clearpage
\onecolumn
\appendix

\input{sections/Appendix}

\end{document}

%% file: sections/abstract.tex
\begin{abstract}

We revisit the common practice of evaluating adaptation of Online Continual Learning (OCL) algorithms through the metric of \textit{online accuracy}, which measures the accuracy of the model on the immediate next few samples. However, we show that this metric is unreliable, as even vacuous blind classifiers, which do not use input images for prediction, can achieve unrealistically high online accuracy by exploiting spurious label correlations in the data stream. Our study reveals that existing OCL algorithms can also achieve high online accuracy, but perform poorly in retaining useful information, suggesting that they unintentionally learn spurious label correlations. To address this issue, we propose a novel metric for measuring adaptation based on the accuracy on the \textit{near-future} samples, where spurious correlations are removed. We benchmark existing OCL approaches using our proposed metric on large-scale datasets under various computational budgets and find that better generalization can be achieved by retaining and reusing past seen information. We believe that our proposed metric can aid in the development of truly adaptive OCL methods. We provide code to reproduce our results at \url{https://github.com/drimpossible/EvalOCL}.

\end{abstract}

%% file: sections/introduction.tex
\vspace{-0.30cm}
\section{Introduction}
\label{introduction}

 The need for learning on continuously changing data streams has led to a proliferation of research in \textit{Online Continual Learning (OCL)} literature \cite{lopez2017gradient,chaudhry2018riemannian}. The primary aim of OCL algorithms is to enable deep models to continuously adapt to new data distributions without compromising the accumulated knowledge. However, we argue that current metrics have deviated significantly from measuring true adaptation capabilities. The majority of recent works on OCL \cite{cai2021online,ghunaim2023real} measure the adaptation of OCL algorithms using the metric of \textit{online accuracy}, defined as the accuracy of a model on the immediate incoming samples. This evaluation practice resulted in the neglect in measuring the capacity of models to retain previous knowledge. As a consequence, several OCL algorithms achieve high online accuracy \cite{cai2021online} but poor performance on other important metrics like \textit{information retention} which measure catastrophic forgetting of past seen data. Poor information retention performance has been attributed to an inherent stability-plasticity trade-off between learning new concepts and remembering old concepts \cite{cai2022improving}. However, the ability to preserve information and representations of past data, such as images of a class like the Eiffel Tower, should be crucial for accurately predicting future images of the same class, even if they were taken under novel weather conditions or camera poses. Hence, we find this justification to be unsatisfactory.

\begin{figure}[t]
   \centering
   \vspace{-0.25cm}
   \includegraphics[width=0.8\columnwidth]{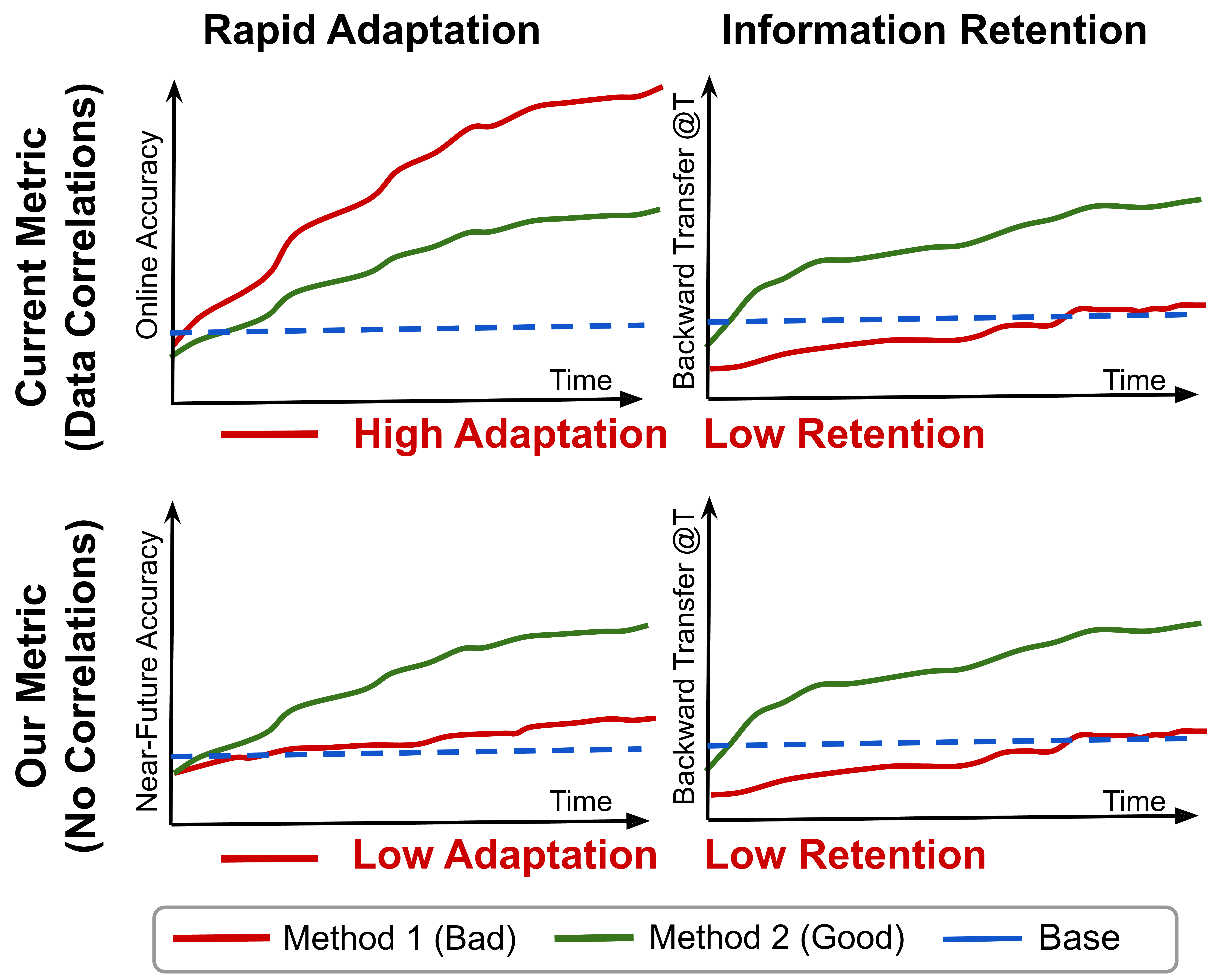}
    \caption{\textbf{Effect of Spurious Correlations.} Current methods in OCL perform well on rapid adaptation but suffer from abysmal information retention (top). We show that these trends are due to idiosyncrasies in the data stream such as label correlations that favor methods that inadvertently overfit to the latest data. Using our proposed metric that removes these correlations (bottom), different sets of methods achieve good performance, with high information retention being key for achieving high adaptability.}
    \vspace{-0.5cm}
    \label{fig:pullfig}
\end{figure}

 In this paper, we demonstrate that the conventional metric of measuring online accuracy may lead to misleading conclusions about the adaptation capabilities of OCL algorithms due to the idiosyncrasies of the data stream. To support our claim, we first demonstrate that a simple blind classifier that relies solely on spurious label correlations in the stream achieves unrealistically high online accuracy on large-scale OCL datasets. Furthermore, we show that this behaviour can be exhibited by deep OCL algorithms, where we propose \textit{OverAdapt} which achieves state-of-the-art performance in online accuracy despite the very poor performance in information retention. We confirm the findings on two large-scale OCL datasets collected from different sources. This indicates that the problem is not limited to a specific dataset, but may be a general concern.
\vspace{0.1cm}

 To overcome this issue, we introduce a new metric called \textit{near-future accuracy} that measures the adaptation capabilities of OCL algorithms by evaluating the accuracy on samples after a shift of $\mathcal{S}$ steps. We choose the minimum shift $\mathcal{S}$ which ensures that the samples no longer have spurious label correlations with the current training data. Choosing the minimum value helps minimize any distribution drift between the training and evaluation samples. The case $\mathcal{S}=0$ recovers the online accuracy metric in case of no spurious correlations. We observe a significant drop in performance on this metric compared to their online accuracy with several state-of-the-art OCL algorithms. This suggests that these algorithms inadvertently rely on label correlations in the stream for predictions. On the contrary, we find that algorithms designed primarily to improve information retention perform exceptionally well in our evaluation strategy. This suggests that a better generalization can be achieved by effectively retaining and reusing past information. Figure \ref{fig:pullfig} illustrates that the best algorithms are those that have little discrepancy between information retention and adaptation. These results challenge current beliefs and demonstrate the importance of using a more precise measure of adaptation in evaluating OCL algorithms.
\vspace{0.1cm}

\noindent Overall, our contributions can be summarized as follows:

\begin{itemize}[leftmargin=*]
\itemsep0em
\item We demonstrate that popular metrics like online accuracy can be unreliable in measuring rapid adaptation due to the presence of spurious label correlations in OCL datasets. 
\item We introduce a novel baseline, OverAdapt, which achieves high online accuracy by relying on label correlations, despite forgetting almost all previously seen data.
\item We propose a novel metric that evaluates the adaptation capabilities of OCL algorithms by measuring accuracy on near-future samples. We select the smallest value of $\mathcal{S}$ that removes label correlations and use this metric to evaluate state-of-the-art OCL methods.
\item Our findings show that existing OCL algorithms perform poorly when evaluated using our proposed evaluation strategy. Algorithms which prevent catastrophic forgetting perform significantly better, suggesting that better adaptation performance can be achieved by retaining and reusing past seen information.
\end{itemize}

\noindent Our results challenge the current emphasis on adapting to the latest samples and demonstrate the importance of seeking a more precise measure of adaptation. We note that our contributions are robust, with our findings valid across various OCL approaches, optimization strategies, and large-scale datasets from different sources.

%% file: sections/relatedwork.tex
\section{OCL Evaluation: A Review}
\label{sec:relatedwork}
\vspace{-0.15cm}

\begin{table}[t]
\scriptsize
\centering
\caption{
\textbf{Properties of OCL Benchmarks.} We list benchmarks in the field of OCL along with the works which introduced them. We compare them across six properties: Realistic Data Streams (DO),  No Storage Constraints (SC), Rapid Adaptation (RA), Information Retention (IR), Restricts Computational Budgets (CB), Evaluates on long stream sizes (LB), Leverages  Pretrained Models (PT). Note that \greencheck is better than \redx in all columns.}

\vspace{0.2cm}
\resizebox{\columnwidth}{!}{
\begin{tabular}{llcccccccc} \hline
DataOrder & Benchmark  & DO & SC & RA & IR & CB & LB & PT \\ \hline 
Task-Inc. & A-GEM\cite{chaudhry2018efficient} & \redx  & \redx & \redx & \greencheck & \redx & \greencheck & \redx  \\ \hline
\multirow{2}{*}{Class-Inc.} & 
MIR\cite{aljundi2019online} & \multirow{2}{*}{\normalsize \redx}  & \multirow{2}{*}{\normalsize \redx} & \multirow{2}{*}{\normalsize \redx} & \multirow{2}{*}{\normalsize \greencheck} & \multirow{2}{*}{ \normalsize \redx} &  \multirow{2}{*}{ \normalsize \greencheck} & \multirow{2}{*}{\normalsize \redx} \\
& DER\cite{buzzega2020dark}  &  &  &  & &  &  \\ \hline
\multirow{3}{*}{Blurry}  & GSS\cite{aljundi2019gradient}  &\multirow{3}{*}{\normalsize \redx}  & \multirow{3}{*}{\normalsize \redx} & \multirow{3}{*}{\normalsize \redx} & \multirow{3}{*}{\normalsize \greencheck} &  \multirow{3}{*}{\normalsize \redx} & \multirow{3}{*}{\normalsize \greencheck } & \multirow{3}{*}{\normalsize \redx}\\
& RM \cite{bang2021rainbow} &  &  &  &  & \redx & \\
& CLIB\cite{koh2021online} &  & &  & &  & \\ \hline

\multirow{4}{*}{Natural} & CLEAR\cite{lin2021clear} & \greencheck & \redx & \greencheck & \greencheck & \redx & \redx & \greencheck\\
& BudgetCL\cite{prabhu2023computationally} & \redx & \greencheck & \redx & \greencheck & \greencheck & \redx & \greencheck  \\
& CLOC\cite{cai2021online} & \greencheck & \redx & \greencheck & \redx & \redx & \greencheck & \redx \\ 
&  DelayOCL\cite{ghunaim2023real} & \greencheck & \redx & \greencheck  & \redx & \greencheck & \greencheck & \greencheck \\
& ACM\cite{prabhu2023online} & \greencheck & \greencheck & \greencheck & \greencheck & \greencheck & \greencheck & \greencheck \\ \hline
\end{tabular}}
\vspace{-0.2cm}
\label{table:cost_intro}
\end{table}

We start with an overview of benchmarks in OCL across different properties of interest. For a more detailed review of the literature, we refer the reader to \cite{Wang2023ACS}.

\vspace{-0.3cm}
\paragraph{1. Realistic Data Streams (DO).} The OCL community has been moving towards increasingly realistic benchmarks in terms of data ordering and task setup. Early works \cite{lopez2017gradient,chaudhry2018efficient} assumed access to which subset of classes a test sample is from (task-incremental setup). Subsequent work removed this constraint, requiring models to predict across all seen classes so far (class-incremental setup) \cite{aljundi2019gradient, aljundi2019online}. However, these works used incoming samples from disjoint data distributions with artificial task boundaries, resulting in unrealistic ordering. Recent works \cite{bang2021rainbow,koh2021online,prabhu2023computationally} improved the ordering by mixing samples from different disjoint tasks. Latest works \cite{cai2021online,prabhu2023online,lin2021clear} eliminate the need for generating artificial data streams simply by using real-world timestamps to order the data stream. Our work focuses on these real-world data streams on large-scale datasets. %

\vspace{-0.3cm}
\paragraph{2. Towards No Storage Constraints (SC).} The majority of prior benchmarks on continual learning constrain the problem by prohibiting access to previously received data \cite{li2017learning, kirkpatrick2017overcoming}, with only a small portion of data allowed to be stored in memory \cite{lopez2017gradient,chaudhry2018riemannian,buzzega2020dark,aljundi2019gradient,koh2021online,lin2021clear}. This constraint is often justified by two reasons: (i) storage space is expensive and (ii) access to previous data is prohibited due to privacy constraints. However, recent work \cite{prabhu2023online,prabhu2023computationally} shows critical shortcomings of these justifications. They show that (a) storage costs are negligible compared to the computational costs of training a model, and (b) simply restricting access to previous data does not address privacy considerations, as samples can be reconstructed from model weights \cite{haim2022reconstructing} or detectably change the model output \cite{shokri2017membership}. Interestingly, Goel \etal \cite{goel2022evaluating} shows that removing data from the model can be done by catastrophic forgetting, which is antithetical to the objective of OCL approaches. Hence, we do not impose any memory constraints on our evaluation.

\vspace{-0.3cm}
\paragraph{3. Towards Better Information Retention (IR).} Interestingly, most of the previous benchmarks in OCL focus on preventing catastrophic forgetting of previously learned information, rather than evaluating the ability of models to rapidly learn new concepts \cite{lopez2017gradient,rebuffi2017icarl,chaudhry2018efficient,bang2021rainbow,aljundi2019gradient,lin2021clear}. This emphasis on information retention can be attributed to disjoint set-based data ordering, which often involves little to no change in distribution for most of the stream, making it difficult to evaluate rapid adaptation. However, recent works have shifted their focus to rapid adaptation performance in real-world ordered data streams \cite{cai2021online, ghunaim2023real}. In our benchmark, our objective is to maintain an additional high degree of retained knowledge from past data as demonstrated in \cite{prabhu2023online, cai2022improving}.

\vspace{-0.3cm}
\paragraph{4. Towards Enabling Rapid Adaptation (RA).} The goal of \textit{online} continual learning algorithms is to rapidly adapt to incoming data from changing distributions. Similarly to past work \cite{prabhu2023online, cai2021online, ghunaim2023real}, we place a strong emphasis on achieving this goal with better metrics.

\vspace{-0.3cm}
\paragraph{5. Towards Computational Budgets (CB).} Continual learning without memory constraints is primarily a problem of achieving high performance with computational efficiency, as retraining from scratch is an ideal solution to the problem. This setting addresses the real-world problem of reducing the cost of updating a given model with incoming data from a stream \cite{prabhu2023online}. We evaluate OCL approaches under the maximum computational budget for fairness \cite{ghunaim2023real}, testing across two different budgets.

\vspace{-0.3cm}
\paragraph{6. Towards Large-Scale OCL (LB).} OCL benchmarks have historically focused on learning over samples incoming from a stream over long data streams since the development of GEM \cite{lopez2017gradient}. Recent OCL benchmarks \cite{aljundi2019gradient, koh2021online, prabhu2023online} have preserved this useful characteristic, while scaling up to larger and more complex data streams. In this work, we use two large-scale OCL benchmarks: \textit{Continual Google LandMarks (CGLM)} \cite{prabhu2023online} and \textit{Continual LOCalization (CLOC)} \cite{cai2021online}. CGLM is a landmark classification dataset that consists of a long-tailed distribution with 10,788 classes that simulate images that arrive on a Wikimedia Commons server. CLOC is a dataset focused on geolocation at scale consisting of 713 classes with 39M images simulating images arriving on a Flickr server. As these datasets come from two different sources, the issue of label correlation that we observe is likely to be a repeating pattern across future OCL datasets.

\vspace{-0.3cm}
\paragraph{7. Leveraging Pretraining for Effective OCL (PT).} Traditional methods in online continual learning start with randomly initialized models \cite{lopez2017gradient, aljundi2019online}. However, continual learning approaches can leverage the abundance of large-scale pretrained models to improve computational efficiency \cite{ostapenko2022continual, wu2022class, prabhu2023online}. We focus on an evaluation setting that
starts with ImageNet1K pretrained models to allow approaches to leverage pretrained information.

\begin{figure*}[t!]
\vspace*{-0.5cm}
\centering
\includegraphics[width=0.75\linewidth]{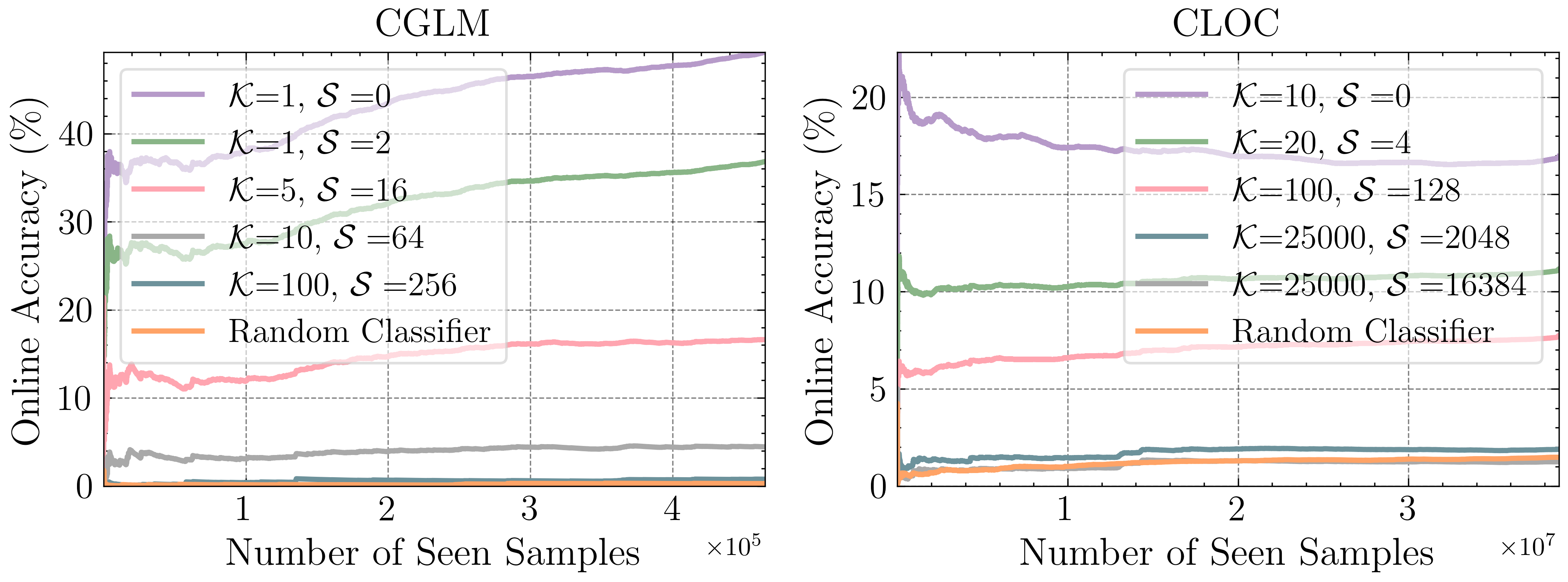}        \caption{\textbf{Blind Classifier.} Performance of a blind classifier on the CGLM (left) and CLOC (right) datasets across varying shift $\mathcal{S}$ towards future for selecting evaluation samples alongside the optimal context window size $k$ for the selected shift. (Section \ref{sec:blind}) A blind classifier achieves unrealistically high accuracy on current evaluation of adaptation (dark blue line, $\mathcal{S}$=0) on both datasets despite being a trivial baseline. (Section \ref{sec35}) The accuracy of the blind classifier on incoming samples with increasing shift $\mathcal{S}$ into the future decreases significantly, eventually converging to near-random performance, indicating a lack of label correlations to the current training samples.}
    \vspace{-0.30cm}
\label{fig:blindclassifier}
\end{figure*}

%% file: sections/our_methodology.tex
\section{Correcting Evaluations of Rapid Adaptation}
\label{methodology}
\vspace{-0.15cm}

We delve into our exploration of the limitations of using online accuracy as a metric to measure adaptation to distribution shifts and present our efforts to alleviate the shortcomings of this metric. We start by identifying the label correlations that affect online accuracy, and then we unveil a technique that enables deep networks to artificially achieve high performance on this metric. This highlights the dependence of the networks on these label correlations. Finally, we introduce a novel evaluation metric, near-future accuracy, and a test to detect and remove label correlations allowing for accurate measurement of the effectiveness of OCL methods in rapidly adapting to new distributions.

\subsection{Evaluating OCL: A Problem Formulation}
\label{sec:problemform}
\vspace{-0.15cm}

In traditional OCL setups with real-world stream orders \cite{cai2021online, prabhu2023online}, the learner is fed a continuous stream of training samples over time steps $t \in \{1, 2, \dots, \infty\}$, consisting of an input datapoint $\mathbf{x}_t  \in \mathcal{X}$ and its corresponding label $\mathbf{y}_t \in \mathcal{Y}$, where $(\mathbf{x}_t,\mathbf{y}_t) \sim \mathcal{D}_{j \leq t}$. The distribution $\mathcal{D}_{j}$ can change at any stream timestep. The aim is to train a classifier $f_{\theta_t}: \mathcal{X} \to \mathcal{Y}$ at any given $t$, which accurately maps a new sample $\mathbf{x}$ to its label $\mathbf{y}$ while adapting to the latest information and incorporating it with the historical knowledge acquired from previous data.

To evaluate the adaptation ability of $f_{\theta_{t}}$, online accuracy measures the performance on the next unseen sample, ($\mathbf{x}_{t+1}$, $\mathbf{y}_{t+1}$), \ie, $a_t = \mathbbm{1}\{f_{\theta_{t}}(\mathbf{x}_{t+1})= \mathbf{y}_{t+1}\}$. Online accuracy is updated in an online fashion given by $A_t^{RA} = \frac{1}{t} \left( A_{t-1}^{RA} \cdot (t-1) + a_t \right)$. For OCL approaches training deep networks, the running average is updated on a batch $\mathcal{B}$ of the next unseen samples as opposed to a single sample. After the evaluation is carried out, the model $f_{\theta_i}$  is updated using the same samples.

The ability of $f_{\theta_{t}}$ to retain information is measured by evaluating its accuracy on an unseen test stream that is similar to the training stream. In particular, per time step $t$, the model trained at the end of step $t$ is evaluated on all test samples from step $0$ to step $t$. We measure Backward Transfer @ $T$ \cite{cai2021online} where $T$ is the last time step of the stream. That is to say, the model at the last step of the stream is evaluated on an increasing test test from step 0 to the end of the stream.
However, traditional OCL methods targeting natural distribution shifts have neglected this metric \cite{cai2021online, ghunaim2023real}.%

\subsection{Isolating and Quantifying Label Correlations} 
\label{sec:blind}
\vspace{-0.15cm}

\noindent Our objective is to investigate whether using online accuracy to measure adaptation in OCL suffers from drawbacks due to label correlations. To accomplish this, we introduce a very simple algorithm that leverages label correlations and performs well in terms of online accuracy.
\newline
\noindent \textbf{Blind Classifier.} We define a blind classifier, similar to prior works \cite{cai2021online, prabhu2023online}, as a model that predicts the mode of the last $\mathcal{K}$ samples seen without access to the input images. Formally, at a given new step $t$, the blind classifier predicts the class of $\mathbf{x}_t$ as the mode of the last revealed labels $\cup_{j=1,\dots,k} \{\mathbf{y}_{t-j}\}$. The optimal context window size $\mathcal{K}$ is selected by a search over different sizes on the hyperparameter search set.

\noindent\textbf{Results.} We present the performance of the blind classifier on the online accuracy in Figure \ref{fig:blindclassifier}. For this section, we focus solely on measuring online accuracy (indicated by the topmost purple line). Our results indicate that the blind classifier performs remarkably well achieving an average accuracy of 50\% on the CGLM dataset, which consists of 10,788 classes, using only the labels of the last seen training sample ($\mathcal{K}=1$). Moreover, the blind classifier achieves 17\% on the large-scale CLOC dataset consisting of 718 classes with $\mathcal{K}=10$ achieving the best performance. These findings suggest an unusually high degree of label correlation between the past 1-10 samples and the immediate incoming sample in the data stream, especially given that the blind classifier achieves 50\% and 17\% accuracy, while the random baseline achieves 0.01\% and 2\% on CGLM and CLOC datasets, respectively.

\noindent \textbf{Conclusion.} %
Despite having never processed any input images, the blind classifier achieves a remarkably high online accuracy. This is surprising because it was expected to perform no better than a random classifier. However, it achieves this result by exploiting the label correlations present in OCL datasets. This is a critical drawback in using online accuracy to measure adaptation. In the following sections, we will delve deeper into this issue.

\subsection{Can Deep Networks Learn Label Correlations?}
\vspace{-0.15cm}

\begin{figure*}[t]
\vspace*{-1cm}
    \centering
    \includegraphics[width=0.8\linewidth]{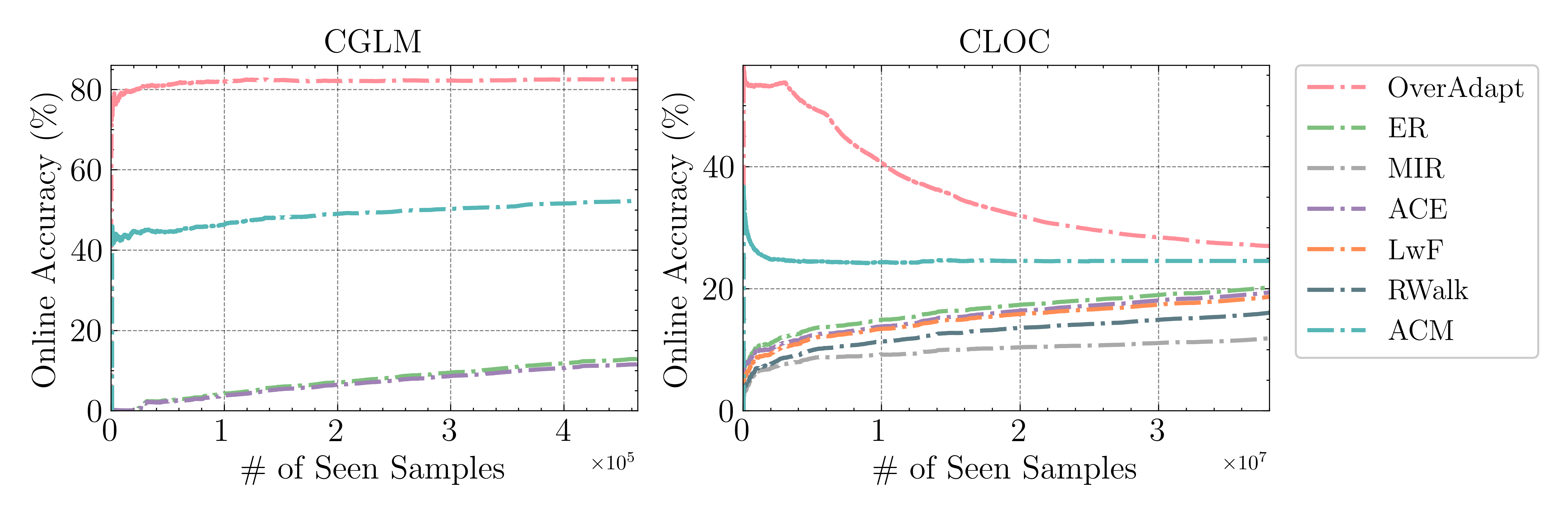}
    \caption{\textbf{OverAdapt Classifier.} We compare the adaptation performance of our OverAdapt classifier on CGLM (left) and CLOC (right) datasets using online accuracy as a measure. The three top performing OCL methods on CLOC are benchmarked on the CGLM dataset. We see that OverAdapt outperforms recent OCL methods by a large margin of 10-40\% in accuracy.}
    \label{fig:overfit}
\end{figure*}

After demonstrating the effectiveness of the blind classifier in exploiting label correlations in the two large-scale datasets used in OCL settings, a pertinent inquiry arises: Can deep networks inadvertently learn these same label correlations leading to unreasonably high online accuracies? This question warrants investigation to determine to what extent deep OCL algorithms can leverage label correlations and the impact this may have on their performance.

\noindent \textbf{An Overfitting-Based Baseline.} One approach for DNNs to exploit label correlations is by overfitting to recent data. We investigate algorithmic choices for designing an OCL method that can overfit the most recent labels in a data stream. Specifically, we make two important design choices for our baseline: (i) we update the model only on recent data to achieve overfitting while sacrificing most of the useful information from the past. To this end, we adopt FIFO sampling, a widely-used technique in OCL, to select batches of samples for training. (ii) We fix the feature representations and train only the last linear layer to prevent degradation of feature representations due to overfitting. To improve the quality of the representations, we use a ResNet50 model pretrained on Instagram1B \cite{mahajan2018exploring}. We refer to this model as \emph{OverAdapt}. OverAdapt is computationally efficient since training the Linear layer over multiple updates is relatively inexpensive. Moreover, we adopt the fast stream evaluation setting \cite{ghunaim2023real}, \ie a setting with a strictly limited computational budget. We set the upper bound for the computation to one gradient update per incoming batch of samples for all methods with no restrictions on storage space. %

\noindent \textbf{Results.} We present the results of the proposed OverAdapt compared to state-of-the-art OCL methods trained using an ImageNet1K pretrained ResNet50 model on the CGLM and CLOC datasets in Figure \ref{fig:overfit}. OverAdapt achieves remarkable performance, surpassing the best-performing method ACM on the CGLM dataset by more than 30\%, and outperforming all but the ACM baseline on the CLOC dataset by 10-35\%. Notably, OverAdapt achieves an impressive 80\% accuracy on the CGLM dataset.%

\noindent \textbf{Conclusion.} After demonstrating that OverAdapt can effectively leverage label correlations from the data stream, we conclude that
traditional design choices like the FIFO buffer in past OCL setups \cite{cai2021online} improved overfitting to the next sample. It is worrying that potential future work or existing works could have inadvertently made similar choices leading to perceived improvement in algorithms when instead it is a subtle case of overfitting to the online accuracy metric. However, we further ensure the validity of our conclusions by investigating whether the effectiveness of this model is due to a good training algorithm or to potential limitations of the evaluation metric. We explore this in the next subsections.

\subsection{Exploring Properties of OverAdapt}
\label{sec34}
\vspace{-0.15cm}

We investigate the reasons behind the impressive performance of our proposed OverAdapt.

\noindent \textbf{Sensitivity Analysis.} We start with the sensitivity study of OverAdapt by analyzing the impact of modifying its two key components: (i) FC-Only training with large-data pretrained initialization and (ii) FIFO Sampling. To study the effect of FC-Only training, we replace it with full-model training using an ImageNet1K-pretrained ResNet50 which has been common in prior OCL work \cite{ghunaim2023real}. We also analyze the effect of FIFO sample selection by comparing it with uniform sampling, which has been shown to be a simple but effective strategy in limited-compute regimes \cite{prabhu2023computationally}. We evaluate their performance in terms of online accuracy and information retention at time $T$ after training on the entire stream following prior art \cite{cai2021online} as discussed in Section \ref{sec:problemform}.%

\begin{figure*}
\vspace*{-1cm}
    \centering
    \includegraphics[width=0.75\textwidth]{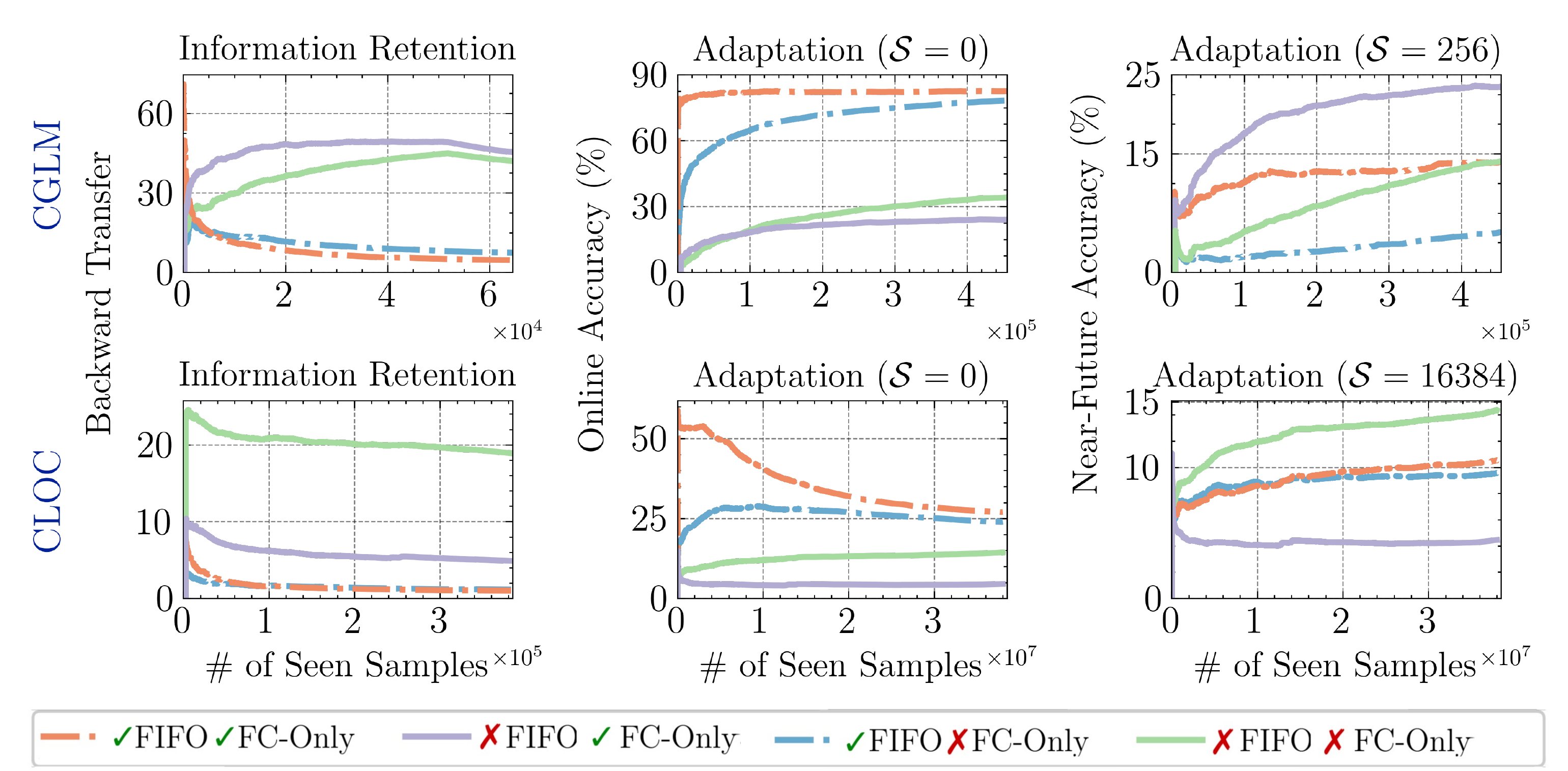}\caption{\textbf{Sensitivity Analysis of OverAdapt Classifier.} We illustrate the performance of various components of our OverAdapt classifier on CGLM (top) and CLOC (bottom) datasets measuring: (Section \ref{sec34}) information retention (left) and rapid adaptation capability in terms of  online accuracy ($\mathcal{S}=0$, center) (Section \ref{sec35}) rapid adaptation performance in terms of online accuracy ($\mathcal{S}=0$, center) with our proposed near-future accuracy ($\mathcal{S}=256$ in CGLM and , $\mathcal{S}=16384$ in CLOC). Note that information retention (left) is independent of the shift $\mathcal{S}$. The results demonstrate that introducing the shift removes label correlations, leading to dramatic loss in  adaptation performance for our OverAdapt method while the uniform baseline stays unaffected.}
\label{fig:overfit_ablation}
\end{figure*}

\noindent \textbf{Results.} Our results are shown in Figure \ref{fig:overfit_ablation} (left \& center).

\textit{\textbf{OverAdapt.}} Strikingly, our analysis reveals that \textcolor{C1}{\textbf{OverAdapt}} (\greencheck FIFO \greencheck FC-Only) achieves remarkable performance in rapid adaptation as measured by online accuracy achieving \textcolor{C1}{\textbf{82\%}} and \textcolor{C1}{\textbf{27\%}} on CGLM and CLOC, respectively. However, this seemingly impressive performance is coupled with an abysmally low information retention when measured by Backward Transfer with \textcolor{C1}{\textbf{OverAdapt}} performing with 5\% and 1\% accuracy on CGLM and CLOC, respectively. These results demonstrate that \textcolor{C1}{\textbf{OverAdapt}} is not a superior OCL algorithm but rather a product of a flawed rapid adaptation evaluation. %

\textit{\textbf{FIFO Sampling.}} Our analysis comparing sampling strategies reveals that models trained using the FIFO sample selection strategy with both \textcolor{C1}{\textbf{FC-Only training}} (\greencheck FIFO \greencheck FC-Only) and \textcolor{C2}{\textbf{full-model training}} (\greencheck FIFO \redx FC-Only) achieve significantly higher accuracy of \textcolor{C1}{\textbf{82}}/\textcolor{C2}{\textbf{78}}\% and \textcolor{C1}{\textbf{27}}/\textcolor{C2}{\textbf{24}}\% on CGLM and CLOC, respectively, on online accuracy metric compared to those trained using uniform sampling-based with \textcolor{C3}{\textbf{FC-Only training}} (\redx FIFO \greencheck FC-Only) and \textcolor{C4}{\textbf{full-model training}} (\redx FIFO \redx FC-Only).
This clearly indicates that FIFO sampling is primarily responsible for the ability to leverage the latest label correlations. However, models trained with FIFO sampling achieve significantly lower accuracy of \textcolor{C1}{\textbf{5}}/\textcolor{C2}{\textbf{7}}\% and \textcolor{C1}{\textbf{1}}/\textcolor{C2}{\textbf{1}}\% on CGLM and CLOC, respectively, in terms of information retention as measured by backward transfer in the last timestep, performing close to a random classifier. This highlights their poor representation quality and confirms that models using the FIFO strategy simply overfit to the latest samples.

\textit{\textbf{FC-Only Training.}} Comparing \textcolor{C1}{\textbf{FC-Only training}} (\greencheck FIFO \greencheck FC-Only) to \textbf{\textcolor{C2}{full-model training}} (\greencheck FIFO \redx FC-Only) under FIFO sampling, we observe that \textcolor{C1}{\textbf{FC-Only training}} achieves a higher accuracy of \textcolor{C1}{\textbf{82}}\% and \textcolor{C1}{\textbf{27}}\% on CGLM and CLOC, respectively, in online accuracy. However, for information retention, both models perform poorly due to the impact of FIFO sampling. In contrast \textcolor{C4}{\textbf{full-model training}} under uniform sampling (\redx FIFO \redx FC-Only) achieved a higher accuracy of \textcolor{C4}{\textbf{34}}\% and \textcolor{C4}{\textbf{14}}\%  on CGLM and CLOC, respectively, for online accuracy, compared to \textcolor{C3}{\textbf{full-model training}} (\redx FIFO \greencheck FC-Only). Additionally, \textcolor{C4}{\textbf{full-model training}} achieved \textcolor{C4}{\textbf{42}}\% and \textcolor{C4}{\textbf{18}}\% accuracy on CGLM (comparable to \textcolor{C3}{\textbf{full-model training}} ) and CLOC (higher than \textcolor{C3}{\textbf{full-model training}}), respectively, in information retention. In general, it is not conclusive whether FC-Only training could substitute full model training, as it performed significantly better in some cases and significantly worse in others. %

\noindent \textbf{Conclusion.} Our experiments highlighted critical drawbacks in assessing the rapid adaptation of deep models using online accuracy, due to spurious label correlations in the data stream.
Sensitivity analysis revealed that both FIFO sampling and FC-Only training are critical components that helped OverAdapt leverage label correlations where FIFO sampling being the primary contributor. Furthermore, our findings suggest that information retention is a useful indicator for identifying drawbacks of OCL algorithms, such as inadvertent overfitting. %

\subsection{Removing Correlations to Evaluate Adaptation} \label{sec35}
\vspace{-0.15cm}

\begin{figure}[t!]
    \centering
    \includegraphics[width=0.71\linewidth]{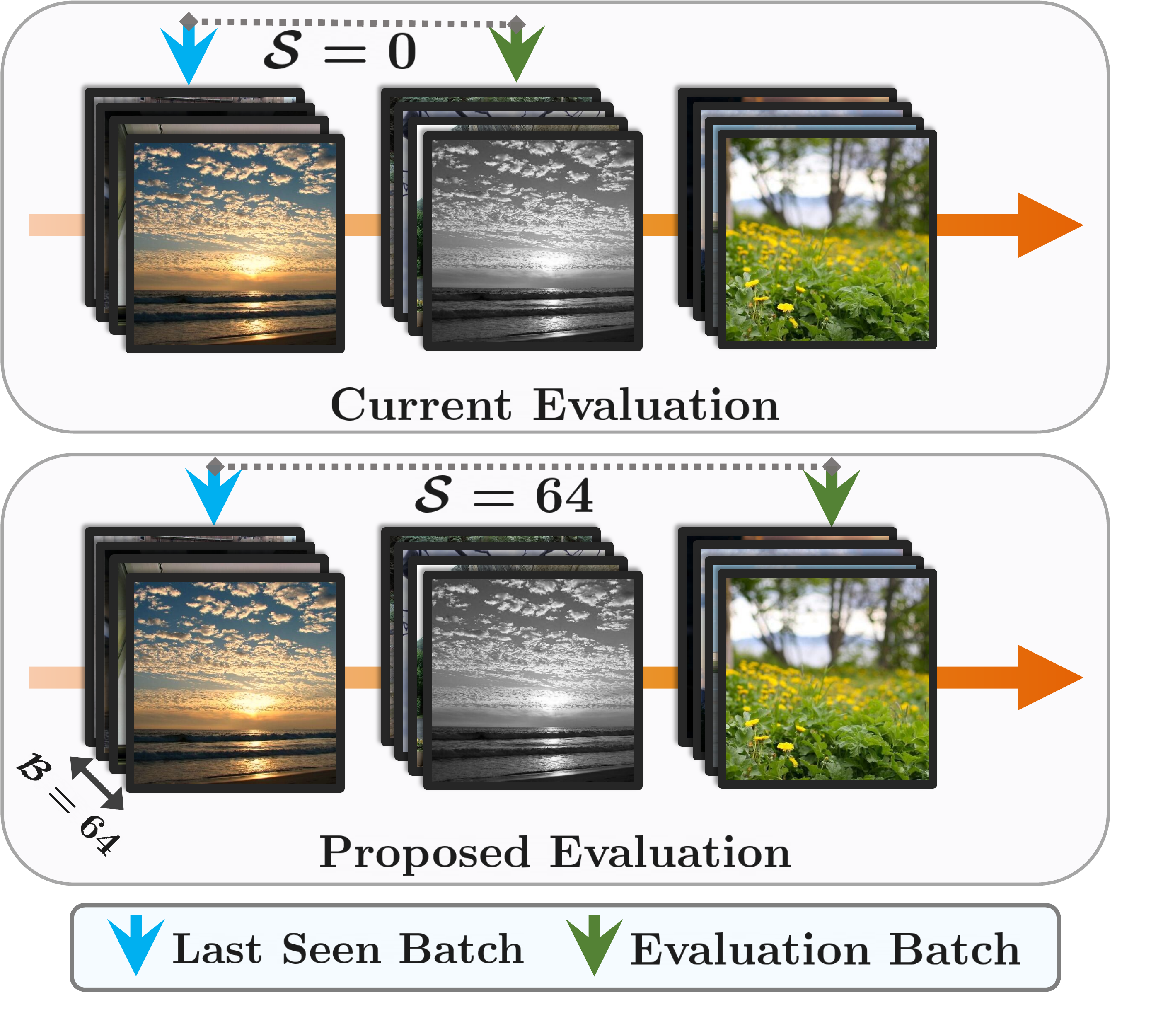}
    \caption{\textbf{Our Evaluation Method.} We compare the samples selected for evaluating adaptation performance by our evaluation method. The blue and green arrows indicate the latest batch received from the stream for learning and the samples selected for evaluation respectively. Current evaluation proposes to compute accuracy on the immediate incoming samples ($\mathcal{S}=0$), \ie online accuracy, while we propose to evaluate on the closest future samples ($\mathcal{S}=64$), \ie near-future accuracy, which is free from spurious correlations with the seen batch of images.}
    \vspace{-0.50cm}
    \label{fig:shiftbench}
\end{figure}

To address the limitations of online accuracy, we propose a new metric, near-future accuracy, which involves evaluating the model on near-future samples rather than immediate next samples. This can mitigate the adverse effects of label correlations on the performance by obtaining a more accurate assessment of the adaptation ability of OCL algorithms.

\noindent\textbf{Formulation of Proposed Evaluation.} As illustrated in Figure \ref{fig:shiftbench}, we introduce a shift $\mathcal{S}$ that represents the number of samples in the future, after which we evaluate the performance of the model. Instead of evaluating on the immediate next sample ($\mathbf{x}_{t+1}$, $\mathbf{y}_{t+1}$), we use the next near-future sample after the smallest shift $\mathcal{S}$, namely ($\mathbf{x}_{t+1+\mathcal{S}}$, $\mathbf{y}_{t+1+\mathcal{S}}$), so that it has no label correlation with them. We select the smallest shift $\mathcal{S}$ instead of an arbitrarily large shift $\mathcal{S}$ so that the test sample comes from the same distribution as the latest seen train samples. We calculate the online accuracy as before, where $\hat{a}_t$ is the running average of the accuracy calculated by first calculating the accuracy $a_t = \mathbbm{1}\{f_{\theta_{t}}(\mathbf{x}_{t+1+\mathcal{S}})= \mathbf{y}_{t+1+\mathcal{S}}\}$ and then updating the running average using the formula $ A_{t}^{RA} = \frac{1}{t} \left(  A_{t-1}^{RA} \cdot (t-1) + a_t \right)$.

\noindent\textbf{How To Set the Shift $\mathcal{S}$?} The shift $\mathcal{S}$ is the smallest value in the future that has no label correlation with the last seen train samples. We select the smallest $\mathcal{S}$ so that the blind classifier performs similarly to a random classifier.%
As shown in Figure \ref{fig:blindclassifier} for both CGLM (left) and CLOC (right) datasets, we observe that increasing the shift $\mathcal{S}$ decreases the accuracy of the blind classifier down to a similar performance of a random classifier. In particular, at a shift of $\mathcal{S}=256$ for CGLM and $\mathcal{S}=16384$ for CLOC, the label correlations are effectively eliminated. Therefore, we use these shift values for our proposed evaluation strategy.
We evaluate OverAdapt and its variants on rapid adaptation with our proposed near-future accuracy and compare it with the previously adopted online accuracy ($\mathcal{S} = 0$).

\noindent \textbf{Results.} Figure \ref{fig:overfit_ablation} presents the results of OverAdapt and its variants on rapid adaptation using two metrics: online accuracy (center) and near-future accuracy (right). When studying FIFO sampling with \textcolor{C1}{\textbf{FC-Only training}} (\greencheck FIFO \greencheck FC-Only) and with \textcolor{C2}{\textbf{full-model training}} (\greencheck FIFO \redx FC-Only), we observe a significant drop in the performance of the methods between the two metrics that overfit using the FIFO sampling by about 70\% in CGLM and 20-40\% on CLOC datasets, while the performance of uniform sampling for both \textcolor{C3}{\textbf{FC-Only training}} (\redx FIFO \greencheck FC-Only) and \textcolor{C4}{\textbf{full-model training}} (\redx FIFO \redx FC-Only) remains similar. Interestingly, the trends reverse, with the uniform sampling variants outperforming the FIFO sampling strategy by a large margin. Furthermore, we observe that FC-Only training no longer consistently improves rapid adaptation, proving to be useful in some cases on the new evaluation method.

\noindent\textbf{Conclusion.} %
We propose evaluation using near-future accuracy which removes label correlations. OverAdapt performs poorly after removing label correlations, compared to a uniform baseline, emphasizing the importance of developing OCL algorithms that handle rapid adaptation without overfitting. Additionally, there is no surprising discrepancy between near-future accuracy and information retention across methods, unlike online accuracy.Our strategy allows for better evaluation of OCL algorithms on real-world ordered datasets.

%% file: sections/experiments.tex
\section{OCL Evaluation by Near-Future Accuracy}
\label{experiments}
\vspace{-0.15cm}

In this section, we provide a comprehensive analysis of various OCL approaches under near-future accuracy. We incorporate the latest advancements in setup design into our benchmark and take computational constraints into account for fair comparison. Following the prior art \cite{prabhu2023online, prabhu2023computationally}, our study assumes no memory constraints, but rather sets constraints on computational budget following \cite{ghunaim2023real, prabhu2023online, prabhu2023computationally}. As addressed in prior work, budgeted computation per time step implicitly imposes a limited memory access.

\subsection{Experimental Setup}
\vspace{-0.15cm}

\noindent\textbf{Datasets and Metrics.} We employ two large-scale online continual learning datasets, CGLM with a total of 460K images and CLOC compromising of 39M images in a stream, both of which contain natural distribution shifts \cite{cai2021online}. That is to say, the images are temporally ordered. We measure rapid adaptation performance using the average accuracy on near-future samples, as described in Section \ref{sec35}, with a shift  $\mathcal{S} = 256$ for CGLM and $\mathcal{S} = 16384$ for CLOC. The choice of both was made such that the blind classifier discussed in Section \ref{sec:blind} achieves a random accuracy. Additionally, we measure information retention using the Backward Transfer metric following CLOC \cite{cai2021online}.

\noindent\textbf{Model and Optimization.} In all experiments, we use a ResNet50 with ImageNet1K initialization unless otherwise specified. We adopt the optimization procedure from \cite{ghunaim2023real} and train all models using an SGD optimizer, a fixed learning rate of $0.005$, and a weight decay of $10^{-4}$. %
For both training and evaluation, we batch incoming samples with a batch size $\mathcal{B}$ of $64$ for CGLM \cite{prabhu2023online} and $128$ for CLOC \cite{ghunaim2023real}.

\begin{figure*}[t]
\vspace*{-0.8cm}
    \centering
    \includegraphics[width=0.75\textwidth]{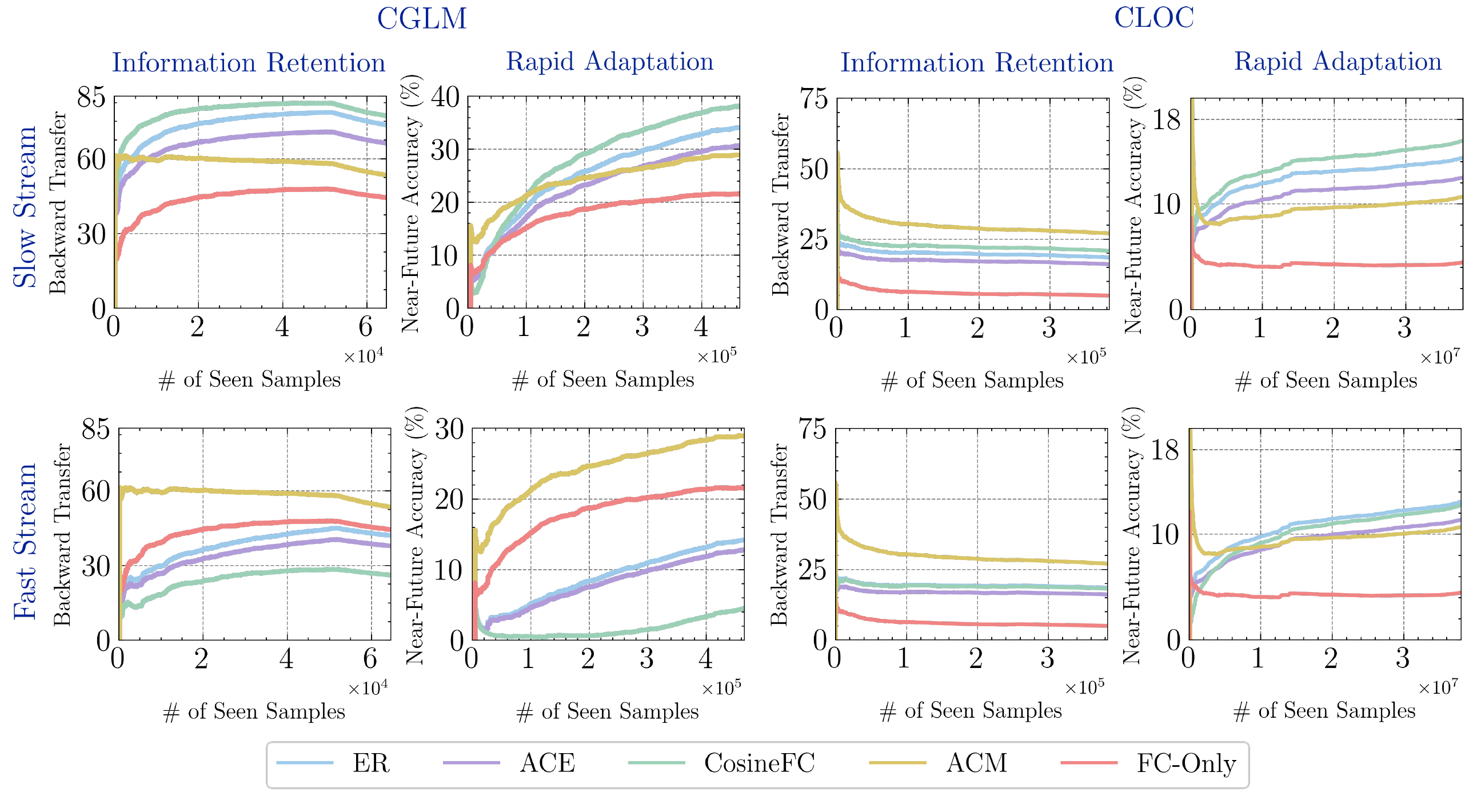}
    \caption{\textbf{Performance of Methods on Near-Future Accuracy}. We evaluate CL methods in slow and fast streams with computational budgets similar to \cite{ghunaim2023real}. Near-future accuracy, which eliminates spurious label correlations, shows that retaining and reusing past information is crucial for better generalization and that algorithms not designed for adaptation in OCL can perform well. Hence retention and adaptation are closely linked.}
    \label{slowfast}
\end{figure*}

\noindent\textbf{Compared OCL Approaches.} %
We evaluate five online continual learning methods using our proposed near-future accuracy. For consistency, we use uniform sampling as the sampling strategy, since it was shown in Section \ref{sec34} that it strikes a good balance between adaptation and retention. Again, in uniform sampling, the training batch is constructed by uniformly and sampling $\mathcal{B}$ samples from all past stored data. The Appendix provides a comprehensive analysis of other sampling techniques.

\textbf{\textit{ER (Replay Only)}\cite{prabhu2023computationally}.} ER (Replay Only) is a simple approach that stores the incoming samples and trains the model on a batch sampled from the set of stored samples. It has demonstrated this to be the leading method in budgeted offline CL. 

\textbf{\textit{FC-Only}.} FC-Only, referred to as OverAdapt (\redx FIFO \greencheck FC-Only) in Section \ref{sec34}, is an adapted version of ER (Replay Only) that uses a ResNet50-I1B feature extractor and trains only the last linear layer. This method has been shown to be computationally efficient \cite{prabhu2023computationally}.

\textbf{\textit{ACE}\cite{caccia2022new}.} We replace the CrossEntropy loss in ER (Replay Only) with the ACE Loss proposed in \ to isolate the effect of ACE loss under the proposed evaluation setting.

\textbf{\textit{ACM}\cite{prabhu2023online}.} ACM extracts features from a pretrained ResNet50 model (trained on ImageNet1k) and continually updates a kNN classifier, with $k=2$ and cosine distance as distance metric, over those features. Additionally, we note that ACM does not train a deep network, thus incurring minimal computational cost. This sets the minimal accuracy requirements for OCL methods.

\textbf{\textit{CosineFC}\cite{hou2019learning}.} 
This approach replaces the linear layer of the ResNet50 model in the ER (Replay Only) baseline with a layer that computes the cosine distance between the weights and the features. We only perform this modification as distillation has been shown to be computationally ineffective \cite{prabhu2023computationally}. 

By comparing these approaches under our proposed near-future accuracy along with information retention, we aim to provide insights into the strengths and limitations of current OCL methods. %

\subsection{Results: Varying Stream Speeds}
\vspace{-0.15cm}

\noindent Following prior work on fair computationally normalized evaluation of online continual learners \cite{ghunaim2023real}, we evaluate the performance of OCL methods in two different time constraints, namely slow stream and fast stream scenarios.

\noindent\textbf{Slow Stream.} Slow stream is the scenario in which the computational budget for learning is not strictly limited. We enforce all methods to perform the equivalent of 10 and 5 model updates per time step on CGLM and CLOC, respectively. Methods that require a significant overhead will thus perform a fewer number of model updates such that it matches the corresponding computational budget of 10 and 5 updates on the respective datasets. Note that ACM requires negligible training cost, and therefore OCL methods must at least outperform ACM to be considered useful.

\noindent\textbf{Results.} In Figure \ref{slowfast} (first row), we present the performance of various methods in terms of near-future accuracy for both CGLM and CLOC datasets. CosineFC, originally proposed for offline continual learning, achieves the highest accuracy with 38\% and 16\% accuracy on CGLM and CLOC, respectively. Furthermore, CosineFC also shows impressive information retention performance, achieving an accuracy of 77\% in CGLM. As expected, FC-Only which only fine-tunes the last linear layer, has the worst retention capabilities due to the restricted access to full-model updates. %

\noindent\textbf{Conclusion.} %
Our evaluation revealed that methods not explicitly designed for adaptation in the OCL setup perform well, indicating that retaining and reusing previously seen information is crucial for achieving better generalization, and that retention and adaptation are closely linked.

\noindent\textbf{Fast Stream.} Fast stream setting refers to the scenario where the computational budget for learning is limited. That is to say, the stream presents samples at a fast rate that provides limited time for learners to train \cite{ghunaim2023real}. To evaluate the performance of OCL methods under this setting, we restrict the methods to only use a computation equivalent to one model update per incoming batch for both datasets.   %

\noindent\textbf{Results.} %
As shown in Figure \ref{slowfast} (second row), our results suggest that the two simplest baselines, namely ACM and FC-Only, achieve the highest accuracy on CGLM with 29\% and 21\%, respectively. On the other hand, ACM achieves the highest information retention of 27\% while simple ER and CosineFC achieve the highest near-future accuracy of around 13\%.
These results indicate that the simple baseline methods outperform the more advanced OCL methods in both adaptation and information retention in this setup.
\textcolor{black}{This highlights that current OCL methods still have a long way to go to achieve satisfactory performance in the fast stream setting where compute is extremely limited.}

\noindent\textbf{Conclusion.} Under our proposed evaluation approach, in which spurious label correlations are removed, the fast stream setting with its limited computational budget shows that simple baselines can outperform more advanced OCL methods in both adaptation and retention. %
This finding emphasizes the need to develop OCL methods that can adapt rapidly in severely budgeted settings.

\subsection{Further Discussion}

Our benchmark of OCL methods on large-scale datasets highlights two key observations. First, the best-performing online adaptation methods often suffer from forgetting due to overfitting to label correlations, challenging the assumption that the performance gap between adaptation and retention is an inherent learning problem \cite{cai2022improving}. Our proposed near-future accuracy helps reveal label correlations and identifies whether the underlying algorithm is inadvertently exploiting them. Second, methods with better retention properties tend to generalize better and achieve improved performance on near-future accuracy in relaxed and restricted computational settings \cite{Gulrajani2020InSO,Wang2021GeneralizingTU}. This aligns well with the domain generalization literature \cite{Gulrajani2020InSO} and suggests that proper training on previous samples can lead to features that generalize well across domain shifts. Finally, we consistently observe that, under limited computational settings, simpler methods outperform their more computationally involved ones.

%% file: sections/conclusion.tex
\section{Conclusion}
\label{conclusion}
\vspace{-0.15cm}

Our work proposes a new measure to evaluate adaptation in OCL algorithms and highlights its limitations compared to the standard online accuracy evaluation. We found that many current OCL methods overfit to idiosyncrasies of the stream rather than genuinely adapting to new data, as revealed by poor information retention. By using our proposed metric, near-future accuracy, we observed that algorithms with good retention also had better generalization and adaptation capabilities. Our proposed evaluation can serve as a sanity check for future OCL algorithms to ensure that they are not learning spurious label correlations.

%% file: sections/Appendix.tex
\section{Effect of Sampling Strategies}

In Figure 4 of the manuscript, we present the baseline ER for two sampling strategies, namely FIFO and uniform. FIFO selects the latest seen samples to train on, whereas Uniform simply randomly and uniformly samples a set of previously seen samples to train on. Mixed sampling is a mix of both FIFO and Uniform, where half of the batch is constructed using FIFO sampling, and the other half using Uniform sampling. 

\subsection{CGLM}

The results for various samplers for both Online Accuracy (Figure \ref{fig:sampler0}) and Near-Future Accuracy (Figure \ref{fig:sampler1}) on CGLM dataset are summarized below:

\begin{figure}[h!]
    \centering
    \includegraphics[width=0.8\textwidth]{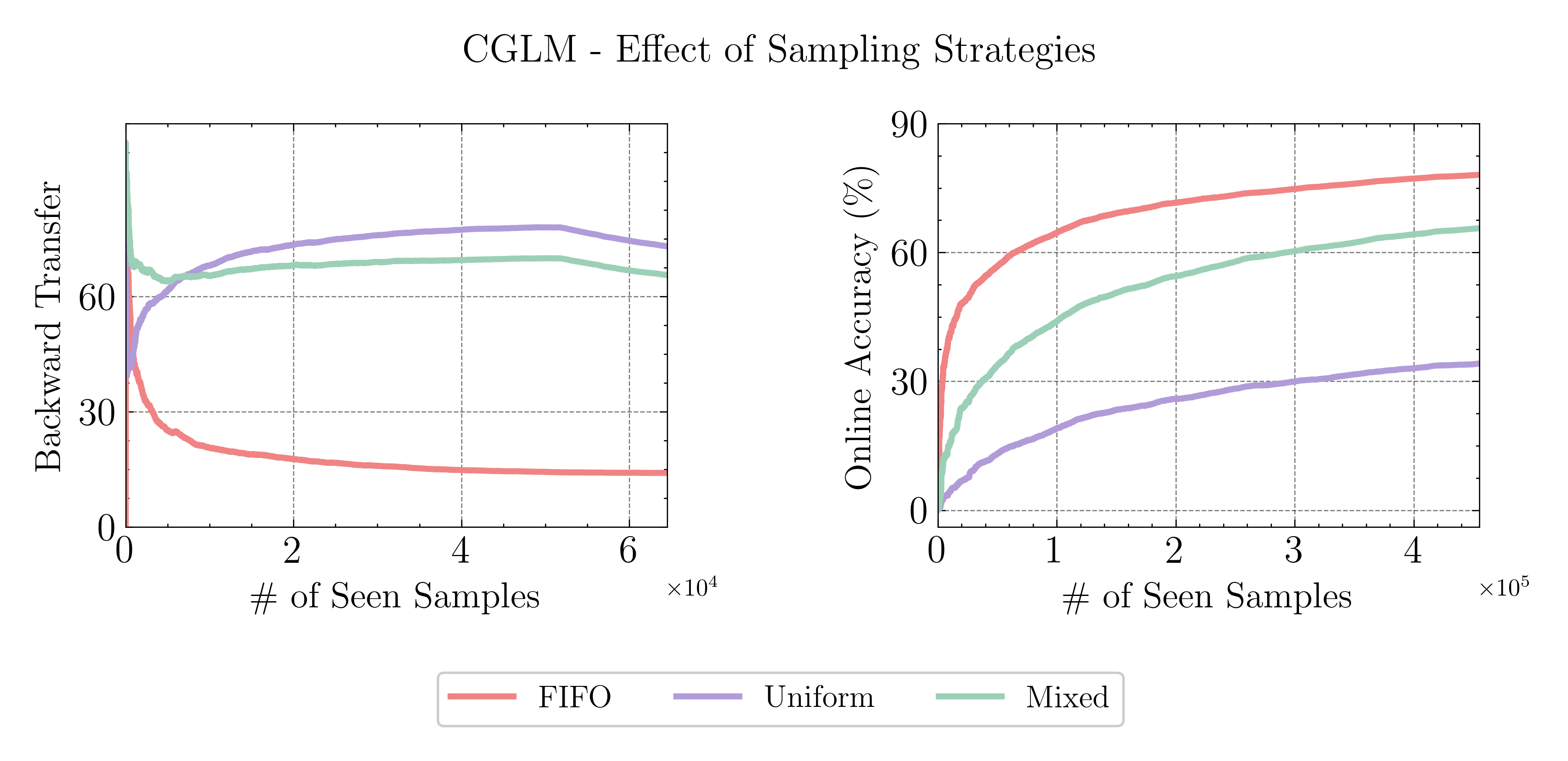}
    \caption{\textbf{Effect of Sampling Strategy on Online Accuracy.} In terms of online accuracy, FIFO sampling which focuses on the latest samples performs best where uniform performs the worst and mixed performs somewhere in the middle. In terms of retetion however, the order is reversed. }
    \label{fig:sampler0}
\end{figure}

\begin{figure}[h!]
    \centering
    \includegraphics[width=0.8\textwidth]{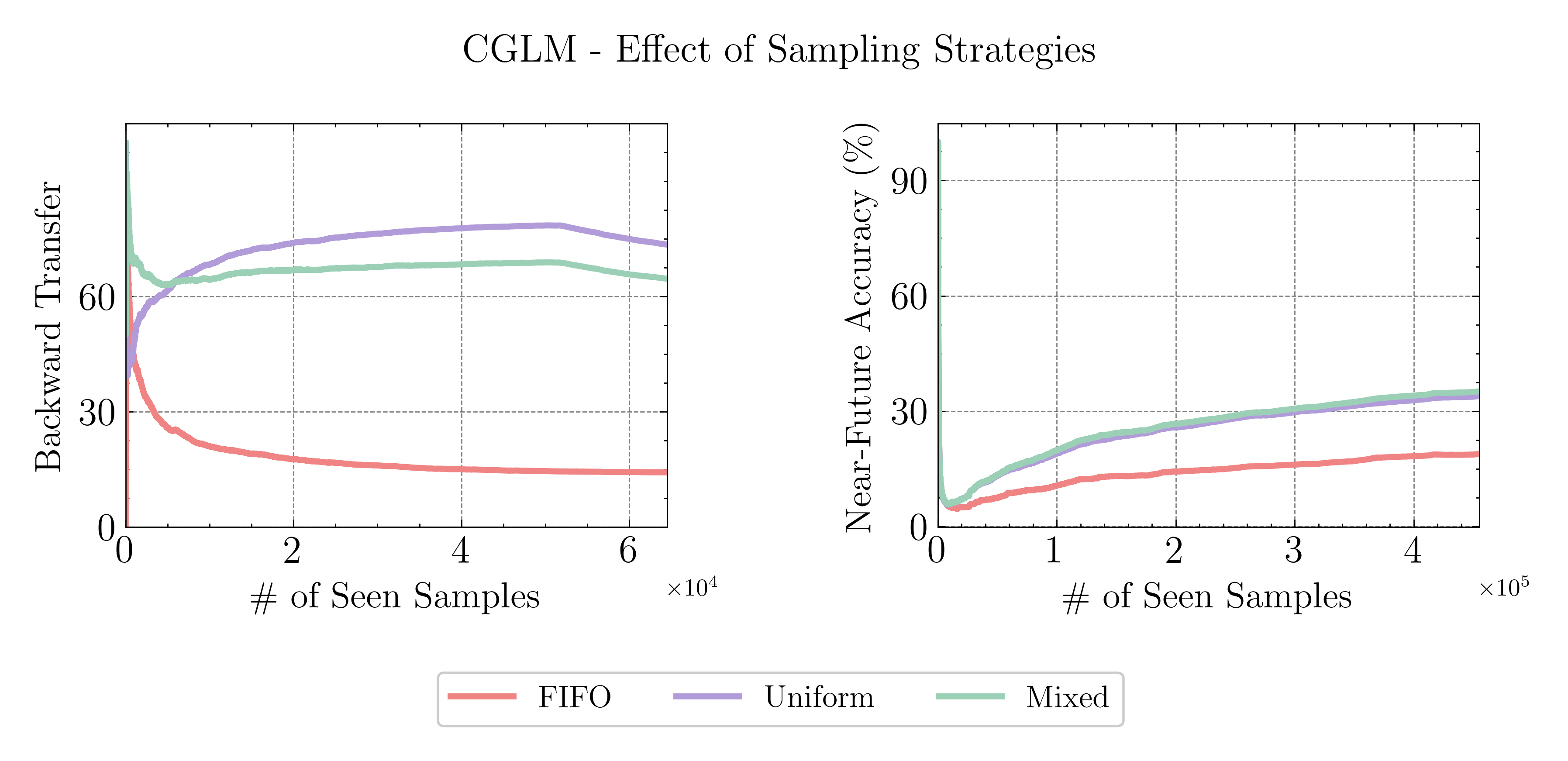}
\caption{\textbf{Effect of Sampling Strategy on Near-Future Accuracy.} In terms of near-future accuracy, Uniform and Mixed sampling perform almost the same, however FIFO sampling is no where close to them. In terms of retention, Uniform still takes the lead.}
\label{fig:sampler1}
\end{figure}

\textbf{Conclusion.} Interestingly, mixed sampling is competitive with uniform sampling in near future accuracy, but performs worse in information retention.

\clearpage 
\subsection{CLOC}

\begin{figure}[h!]
    \centering
    \includegraphics[width=0.8\textwidth]{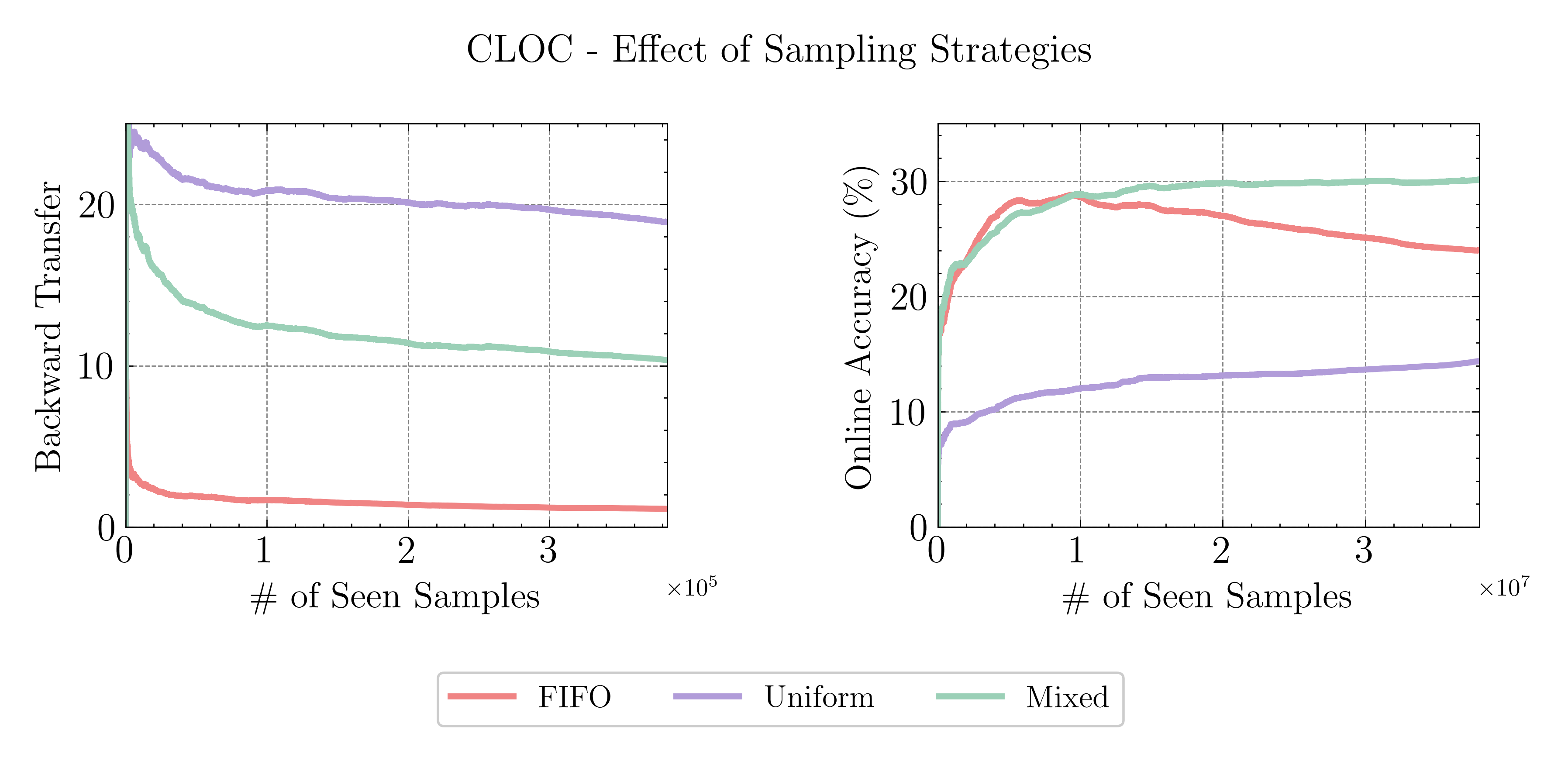}
    \caption{\textbf{Effect of Sampling Strategy on Online Accuracy.} Unlike what was observed for CLGM, for CLOC, the mixed sampling performs the best in terms of online accuracy followed followed by FIFO and then Uniform. However, Uniform still performs the best in terms of information retention with a large gap with a large margin. }
    \label{fig:sampler2}
\end{figure}

\begin{figure}[h!]
    \centering
    \includegraphics[width=0.8\textwidth]{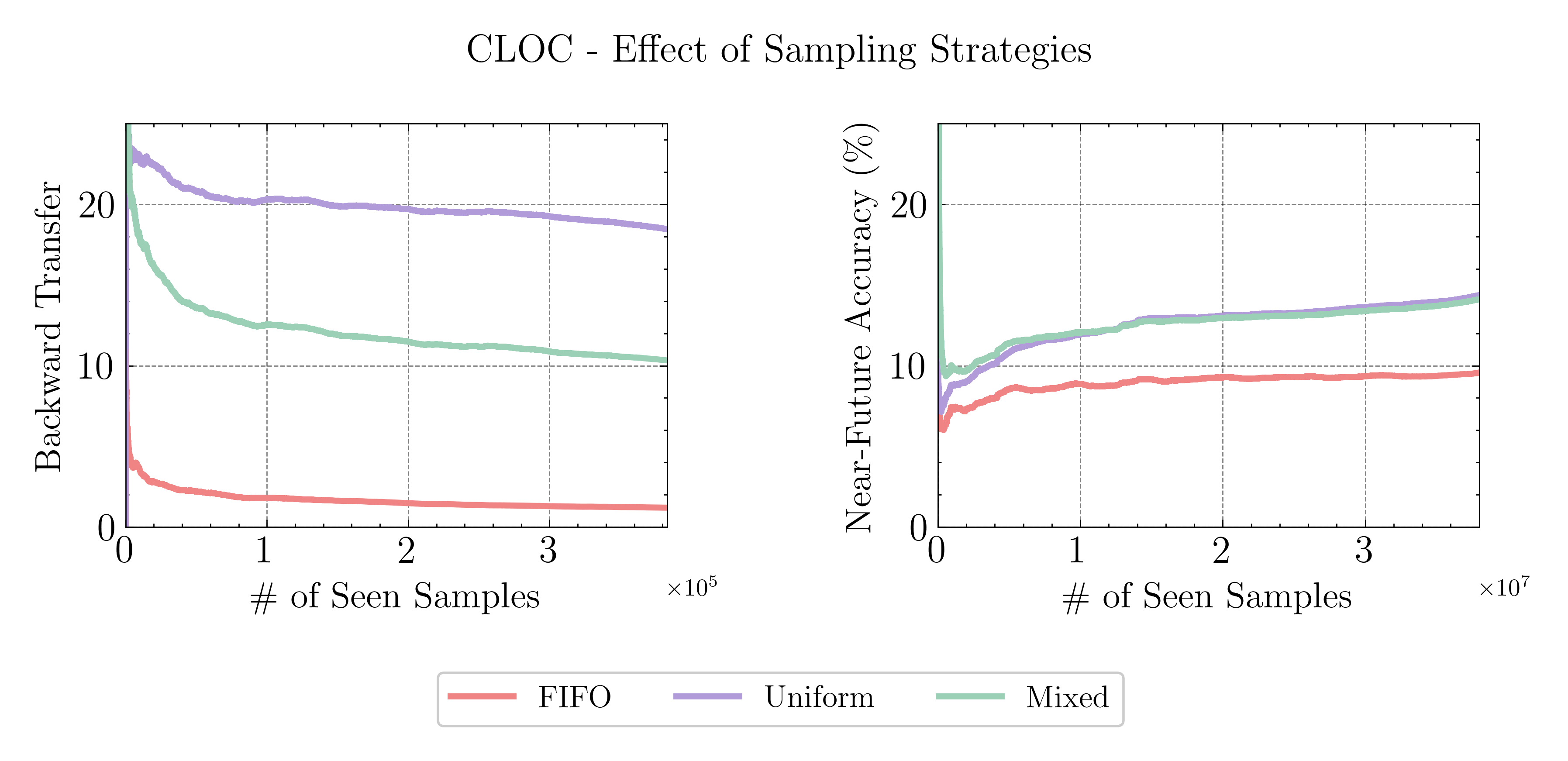}
\caption{\textbf{Effect of Sampling Strategy on Near-Future Accuracy} In terms of near-future accuracy, mixed outperforms other samplers by a significant margin. However, it achieves significantly worse performance compared to uniform sampling in terms of information retention.}
\label{fig:sampler3}
\end{figure}

The results for various samplers for both Online Accuracy (Figure \ref{fig:sampler2}) and near-future accuracy (Figure \ref{fig:sampler3}) on CLOC dataset are summarized below.

\textbf{Conclusion.} Mixed sampling is competitive with uniform sampling in near future accuracy, however, its information retention capabilities are half that of the ER baseline.

\clearpage

\section{Sensitivity Analyis: Learning Rate and Weight Decay}

In OCL, changing hyperparameters across datasets is uncertain as the stream might have a significant distribution shift from the pretrain. Hence, we use the hyperparameters for our model from Ghunaim \etal \cite{ghunaim2023real} for all our experiments. However, how do the selected hyperparameters transfer to CGLM is an interesting question. We demonstrate the sensitivity of the hyperparameters: learning rate and weight decay below:

\subsection{Sensitivity to Weight Decay}

\begin{figure}[h!]
    \centering
    \includegraphics[width=0.75\textwidth]{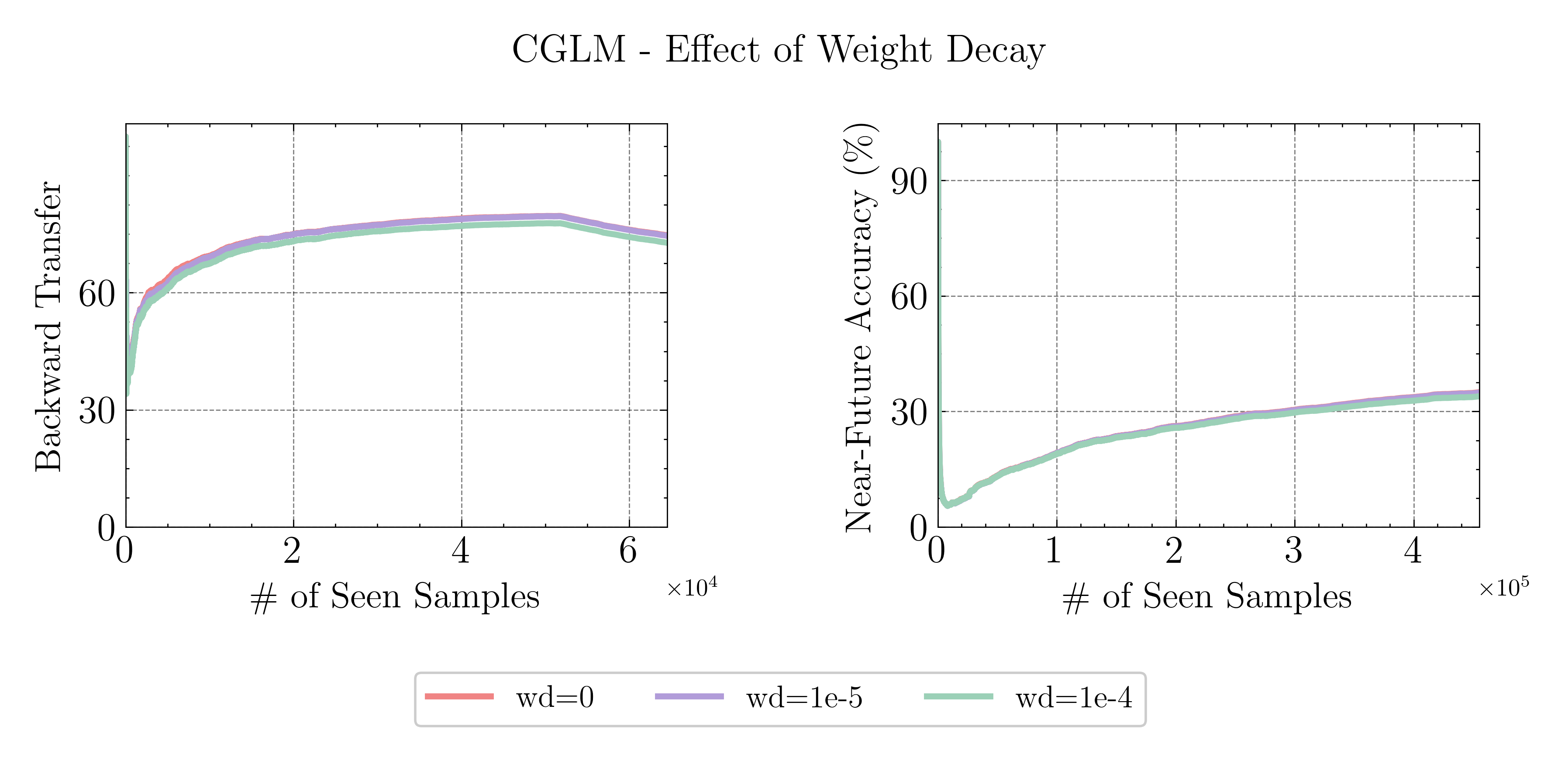}
    \caption{\textbf{Sensitivity to Weight Decay on ER (Replay Only).} Weight decay seems to have minimal effect on both near-future accuracy and backward transfer.}
    \label{fig:wd}
\end{figure}

\textbf{Results and Conclusion.} We present our results in Figure \ref{fig:wd}. We conclude that weight decay has minimal effect on the performance of the ER (Replay Only) method. 

\subsection{Sensitivity to Learning Rate}

\begin{figure}[h!]
    \centering
    \includegraphics[width=0.75\textwidth]{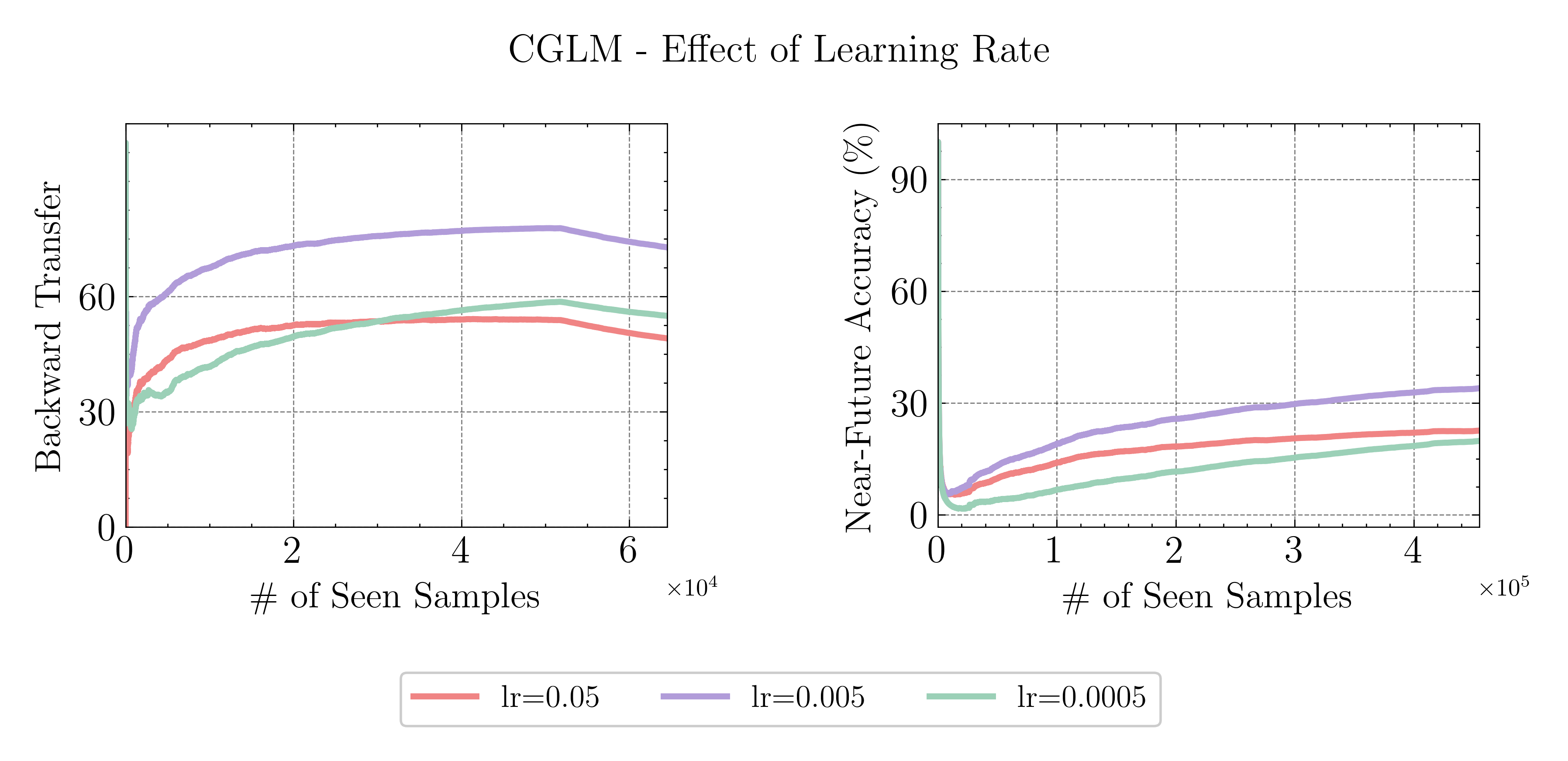}
    \caption{\textbf{Effect of Learning Rate on the ER Baseline on CGLM.} Both near future-accuracy and backward transfer are highly sensitive to the selection of the learning rate. }
    \label{fig:lr}
\end{figure}

\textbf{Results and Conclusion.} We present our results in Figure \ref{fig:lr}. An order magnitude change in learning rate in either direction leads to a decrease in both information retention performance and near-future accuracy. The learning rate of 0.005 transfers well to CGLM dataset in terms of both near-future accuracy and information retention (Backward Transfer).

\clearpage
\section{Limitations}

\textbf{Limitations.} While our work discovers interesting phenomena in OCL, it has important limitations. Firstly, it is unclear why our proposed evaluation approach definitively measures rapid adaptation, and further investigation is needed to determine the problem exists in future OCL scenarios. Secondly, we currently only measure and remove correlations from labels $p(y)$, while covariate correlations $p(X)$ are harder to reliably isolate, requiring further investigation. Lastly, our results suggest that there is still a lot of room for improvement in OCL methods, as we are far from the pareto frontier where information retention and rapid adaptation are at odds.